%% file: main.tex
\documentclass[11pt,a4paper]{article}
\usepackage[hyperref]{main}
\usepackage{times}
\usepackage{latexsym}
\usepackage{enumitem}
\usepackage{url}
\usepackage{graphicx}
\usepackage{amsmath}
\usepackage{amsfonts}
\usepackage{amssymb}
\usepackage{bbm}
\usepackage{booktabs}
\usepackage{xcolor}
\usepackage{siunitx}
\usepackage{pifont}
\usepackage{pbox}
\usepackage{makecell}

\usepackage{multirow}
\usepackage{microtype}

\aclfinalcopy 
\def\aclpaperid{3268} 

\title{Leakage-Adjusted Simulatability: Can Models Generate Non-Trivial Explanations of Their Behavior in Natural Language?}

\author{Peter Hase, Shiyue Zhang, Harry Xie,\and Mohit Bansal \\
  UNC Chapel Hill \\
  \texttt{\{peter, shiyue, fengyu.xie, mbansal\}@cs.unc.edu}\\}

\date{}

\begin{document}
\maketitle

\begin{abstract}

Data collection for natural language (NL) understanding tasks has increasingly included human explanations alongside data points, allowing 
past works to introduce models that both perform a task and generate NL explanations for their outputs. 
Yet to date, model-generated explanations have been evaluated on the basis of surface-level similarities to human explanations, both through automatic metrics like BLEU and human evaluations. 
We argue that these evaluations are insufficient, since they fail to indicate whether explanations support actual model behavior (faithfulness), rather than simply match what a human would say (plausibility). In this work, we address the problem of evaluating explanations from the model \emph{simulatability} perspective. Our contributions are as follows: 
(1) We introduce a \emph{leakage-adjusted simulatability} (LAS) metric for evaluating NL explanations, which measures how well explanations help an observer predict a model's output, while controlling for how explanations can directly leak the output. We use a model as a proxy for a human observer, and validate this choice with two human subject experiments.
(2) Using the CoS-E and e-SNLI datasets, we evaluate two existing generative graphical models and two new approaches; one rationalizing method we introduce achieves roughly human-level LAS scores. 
(3) Lastly, we frame explanation generation as a multi-agent game and optimize explanations for simulatability while penalizing label leakage, which can improve LAS scores.\footnote{We provide code for the experiments in this paper at \url{https://github.com/peterbhase/LAS-NL-Explanations}.}

\end{abstract}

\input{introduction}

\input{related_works}

\input{method}

\input{experiments_and_results}

\section{Conclusion}
We introduce a leakage-adjusted simulatability metric to evaluate the influence of natural language explanations on model simulatability while controlling for explanations leaking the model outputs. We validate our metric with two human subject experiments, and find that: (1) our \textsc{ST-Ra} model attains similar LAS scores to human explanations, (2) rationalizing methods do better than reasoning methods, (3) no statistically significant relationship emerges between simulatability and accuracy, (4) our automatic metric correlates with expert simulation results, (5) the strongest predictor of crowdsourced explanation ratings is whether explanations leak the answer choice, and (6) optimizing explanations for simulatability can improve LAS scores.  

\section*{Acknowledgements}

We thank the reviewers for their helpful feedback. This work was supported by NSF-CAREER Award 1846185, DARPA MCS Grant N66001-19-2-4031, Royster Society PhD Fellowship, Microsoft Investigator Fellowship, and Google and AWS cloud compute awards. The views contained in this article are those of the authors and not of the funding agency.

\bibliography{main}
\bibliographystyle{acl_natbib}

\appendix

\input{appendix}

\end{document}

%% file: introduction.tex
\section{Introduction}

Deep neural models have achieved impressive success in many areas. However, their interpretability and explainability have remained broadly limited. To make neural models more interpretable, previous works have proposed methods for explaining model decisions, e.g., through various feature importance estimates \cite{hendricks2018grounding, ribeiro_why_2016} or model-generated natural language (NL) \cite{hendricks_generating_2016, kim_textual_2018}. Early work on generating NL explanations focused on providing explanations that were both descriptive of an image and discriminative as labels \cite{hendricks_generating_2016}. Since then, a variety of datasets have been collected with free-form human generated explanations accompanying each data point \cite{camburu_e-snli:_2018, kim_textual_2018, zellers_recognition_nodate, wang_does_2019, rajani_explain_2019}. Models have been proposed for these datasets with two aims: (1) to teach models how to explain their own decisions in natural language, by offering demonstrations of humans doing this, and (2) to increase model accuracy on the task, by making use of additional information in human explanations. 

\begin{figure*}[t]
    \centering
    \includegraphics[width=0.8\textwidth]{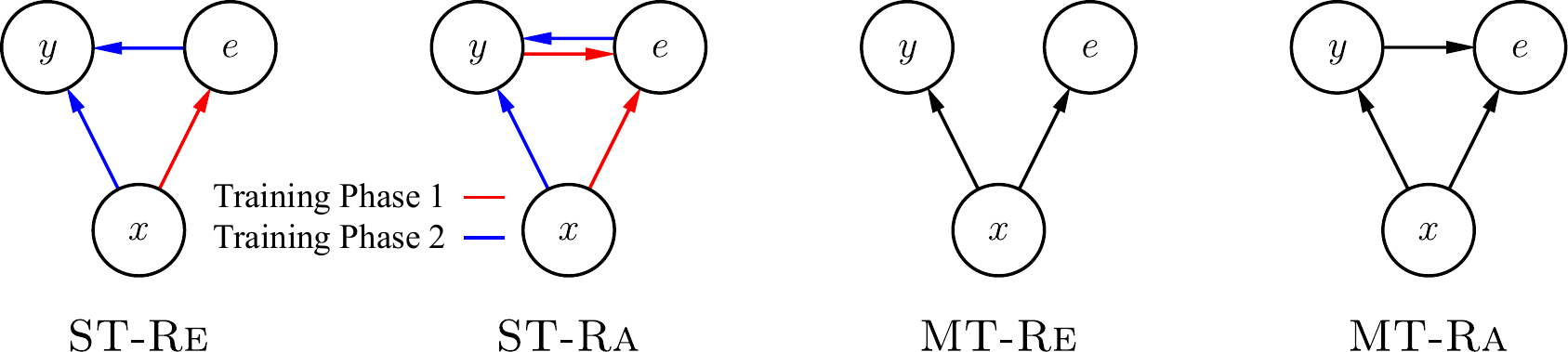}
    \vspace{-3pt}
    \caption{Graphical models representing varying roles of explanations, where the task input is denoted by $x$, task output by $y$, and explanation by $e$.
    We introduce a new rationalizing model, \textsc{ST-Ra}, while also testing a reasoning multi-task model, \textsc{MT-Re}, and two other methods from past works \cite{camburu_e-snli:_2018, rajani_explain_2019}. 
    }
    \label{fig:lead}
    \vspace{-7pt}
\end{figure*}

Past works have proposed varying methods for generating NL explanations, which can be represented by distinct graphical models. 
In our work, we explore four graphical models, shown in Figure~\ref{fig:lead}. Each model generates explanations in either a reasoning (\textsc{Re}) or rationalizing (\textsc{Ra}) mode, where rationalizing models explicitly condition explanations on a label and reasoning models condition only on the input. Approaches further differ by whether they use explanations as inputs to a task model (\textsc{ST}) or as additional supervision in a multi-task framework (\textsc{MT}). Two of these models are drawn from prior works: \textsc{MT-Ra} \cite{camburu_e-snli:_2018} and \textsc{ST-Re} \cite{rajani_explain_2019}. We introduce \textsc{ST-Ra} and also test \textsc{MT-Re} as the reasoning counterpart to \textsc{MT-Ra}. 
To fairly compare the approaches, we implement each graphical model with a state-of-the-art pretrained T5 model \cite{raffel_exploring_2019} (details in Section~\ref{sec:methods_modeling_explanations}). 

Generated explanations have typically been evaluated by automatic measures of similarity with human explanations. Most commonly, phrase-matching metrics such as BLEU \cite{papineni2002bleu} are used. In a few cases, human evaluations have been employed, also primarily to assess the similarity of explanations to what humans would say. On the basis of these evaluations, past works have suggested their models produce ``justifications of its classification decisions" \cite{camburu_e-snli:_2018} and ``explanations to justify its predictions" \cite{rajani_explain_2019}. While useful starting points, we argue that these evaluations are insufficient, because they do not necessarily indicate anything about a model's true internal reasoning. For example, suppose the ground-truth label is A, while a model predicts B; a higher BLEU score will be observed when the model gives an explanation to support human label A, instead of model prediction B. This point is substantiated by \citet{jacovi2020towards}, who advocate for evaluations of explanation \emph{faithfulness} rather than \emph{plausibility}. 

To resolve this evaluation problem, we introduce the \emph{leakage-adjusted simulatability} (LAS) metric, which is better suited for identifying when explanations actually support model behavior. LAS scores combine two key mechanisms: they measure \emph{simulatability}, which reflects how well an observer can use model explanations to predict the model's output, while controlling for explanation leakage, which occurs when explanations directly leak the output. This metric is inspired by prior work on model interpretability \cite{doshi-velez_towards_2017, hase_evaluating_2020}, but to date no simulatability analysis has been carried out for NL explanations. We automate our evaluation by using a pretrained language model as the observer, serving as a proxy for a human.
Using LAS scores, we evaluate model-generated as well as human explanations for  \textsc{CommonSenseQA} (CQA) \cite{talmor_commonsenseqa_2019, rajani_explain_2019} and SNLI \cite{bowman_large_2015, camburu_e-snli:_2018} tasks.
We provide two human evaluations to validate our model-based approach. The first is an expert simulatability evaluation, where we manually play the role of the simulator in our LAS metric computation. The second is a subjective ratings task, where we collect data from Mechanical Turkers. 

Lastly, since we propose a metric for evaluation, the question naturally arises of whether an objective besides standard language modeling is better suited to improving explanations under this metric. While our formulation of LAS is not differentiable, we present a proxy objective that involves using a simulator during training. This training procedure is neatly interpreted as a multi-agent game. Agents share a common objective, which is for the simulator to predict the task model's output using the explanation it receives, but we penalize agents for pursuing the trivial solution, i.e., restating outputs without giving additional information. 

We summarize our key results as follows:
\begin{enumerate}[nosep, wide=0pt, leftmargin=*, after=\strut]
    \item We introduce the LAS score, which captures how explanations improve simulatability while controlling for direct label leakage, and we use it to evaluate four generative models.
    \item We show that our LAS scores provide a deeper understanding of explanation effectiveness than metrics like BLEU and discuss their relationship with our expert simulation analysis and crowdsourced human quality ratings.
    \item We find that our \textsc{ST-Ra} approach achieves nearly human-level LAS scores, and that rationalizing models outperform reasoning models.
    \item We observe no trade-off between interpretability and accuracy, though this also means that existing methods struggle to learn from human explanations.
    \item In a multi-agent game, we show that optimizing explanations for simulatability and penalizing trivial explanations can improve LAS scores in some settings. 
\end{enumerate}

%% file: related_works.tex
\section{Related Work}

\vspace{2pt}
\noindent\textbf{Generating Natural Language Explanations.}
Early work on this topic proposes to generate explanations for images that are descriptive as captions and discriminative as labels \cite{hendricks_generating_2016}.
However, they seek to explain the image's label rather than a classifier's output.
\citet{ling_program_2017} introduce induction approaches for solving math problems and generating explanations of solutions. Two works focus on multi-modal problems, explaining visual question answering \cite{park_multimodal_2018} and self-driving car decisions \cite{kim_textual_2018}. 
A few recent works focus on explanations for language understanding tasks. \citet{camburu_e-snli:_2018} introduce e-SNLI, extending the SNLI dataset \cite{bowman_large_2015} with free-form human explanations, and they provide an LSTM-based model that jointly predicts labels and generates explanations, shown by \textsc{MT-Ra} in Figure~\ref{fig:lead}. \citet{rajani_explain_2019} propose the CoS-E dataset, collecting human explanations for \textsc{CommonSenseQA} \cite{talmor_commonsenseqa_2019}, and they introduce the \textsc{CAGE} model, depicted as \textsc{ST-Re} in Figure~\ref{fig:lead}. 
We build on these works by evaluating both \textsc{ST-Re} and \textsc{MT-Ra} as well as models we introduce, \textsc{ST-Ra} and \textsc{MT-Re}.
We implement each graphical model with strong pretrained-T5 models, and for completeness, we also test methods with GPT2 and BERT (results in Appendix~\ref{appendix:alternative}) \cite{radford_language_nodate, devlin_bert_2019}.

\vspace{2pt}
\noindent\textbf{Evaluating Explanations.}
There is now a wealth of work on evaluating explanations of machine learning models \cite{ribeiro_why_2016, doshi-velez_towards_2017, hooker_benchmark_2019, jacovi2020towards}. For NLP tasks, past works focused on extractive rather than generative explanations \cite{nguyen_comparing_2018, deyoung_eraser_2019}. Such methods extract parts of the model input that are important to the output according to some criterion. However, they are not suited to evaluate NL explanations that are not part of the input, which motivates our new simulatability metric.

Measures of similarity between model-generated and human explanations are used to evaluate nearly every method introduced above, with BLEU being the most common \cite{hendricks_generating_2016, ling_program_2017, park_multimodal_2018, kim_textual_2018, camburu_e-snli:_2018, rajani_explain_2019}. 
In a few cases, human evaluations are employed for similar purposes 
\cite{hendricks_generating_2016, park_multimodal_2018, kim_textual_2018}. 
While these evaluations provide a good starting point, they do not support previous claims that explanations show the reasons for model behavior because they evaluate plausibility rather than faithfulness.
We introduce a leakage-adjusted simulatability metric (LAS) in response to this issue. As observed by \citet{jacovi2020aligning}, faithfulness and simulatability are closely related, but simulatability primarily captures causal attribution of explanations and not necessarily social attribution. 
Simulatability-based evaluations have been conducted before \cite{ribeiro_anchors:_nodate, hase_evaluating_2020}, but we are the first to consider NL explanations and employ model-based controls for label leakage. 
Two contemporaneous works also explore relevant topics. \citet{narang_wt5?!_2020} train a T5 model to generate explanations in a set-up analogous to our \textsc{MT-Ra} setting.
They also notice the shortcomings of BLEU and collect binary human ratings of whether explanations ``support" model outputs. \citet{kumar_NILE_2020} introduce label-specific versions of the method in \citet{rajani_explain_2019}, one of which shares the graphical structure of our \textsc{ST-Ra} model. However, their evaluation focuses on whether humans can recover ground truth labels from generated explanations alone, which they term ``explanation accuracy." 
Given these interesting concurrent works, our contributions are still distinguished by our joint focus on (1) simulatability-based evaluation, (2) controls for explanation label leakage, and (3) comparison of several distinct graphical models.
\begin{figure*}[th]
\centering
\includegraphics[width=.96\textwidth]{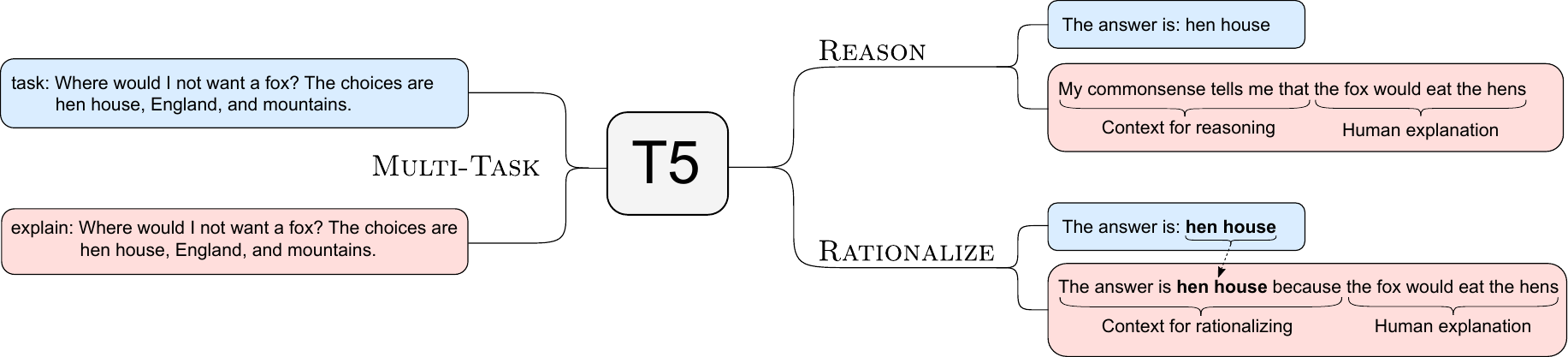}
\vspace{-4pt}
\caption{Inputs and outputs for the T5 Multi-task framework. In the reasoning mode, explanations are not conditioned on the model's prediction, whereas in the rationalizing mode they are dependent on the model output.} 
\label{fig:t5}
\vspace{-2pt}
\end{figure*}

\vspace{2pt}
\noindent\textbf{Multi-Agent Communication.}
The most relevant work to our multi-agent game concerns discrete communication policies with natural language or artificial protocols grounded in NL. \citet{lazaridou_multi-agent_2017} ground a communication protocol in natural language via an auxiliary image classification task. In concurrent work, \citet{lazaridou2020} learn NL protocols for an image-based reference game by pretraining with image captions. While our approach shares the premise that language use is goal-oriented, we optimize full explanations of model outputs rather than descriptions of images in reference games.
Another contemporaneous work optimizes for simulatability in a multi-agent setting, but they use extractive rather than generative explanations \cite{treviso_towards_2020}.

%% file: method.tex
\section{Modeling With Explanations}
\label{sec:methods_modeling_explanations}

In this section, we delineate our baseline model and the four graphical models we study. The graphical models are depicted in Figure~\ref{fig:lead}. 
We also summarize the key features of each approach in Table \ref{table:conditions}. We show examples of task inputs and outputs along with explanations in Table~\ref{tab:exp-examples}. In general, we initialize models from T5-Base, which is a Transformer-based sequence-to-sequence model, pretrained with a large-scale English corpus.

\paragraph{Baseline.} The baseline model simply predicts $y$ given $x$. 
We adopt the approach of \citet{raffel_exploring_2019} for fine-tuning to multiple-choice problems, which is to maximize the likelihood of correct answer tokens conditioned on the task inputs.
To produce predictions, however, we compute a likelihood for each answer choice and select the most likely choice, rather than sampling text. SNLI also fits into this framework by taking the three relations as answer choices.

\paragraph{\textsc{ST-Re}.} \citet{rajani_explain_2019} proposed a Commonsense Auto-Generated Explanation (CAGE) framework for CQA, with a two-phase training procedure: 
first, with human explanations as supervision, a model is trained to generate explanations given task inputs; then generated explanations are supplied with task inputs to a classifier that performs the task. 
We represent this framework in Figure~\ref{fig:lead}, where we term it \textsc{ST-Re} to fit within our data-agnostic model taxonomy. ST stands for serial-task (from the separate training phases) and \textsc{Re} for the reasoning explanation generation. 
While originally composed of GPT and BERT, we implement this approach with two separate T5 models. 

\paragraph{\textsc{ST-Ra}.} We extend the \textsc{ST-Re} approach to operate in a rationalizing mode (shown in Figure \ref{fig:st_ra} in Appendix). Instead of generating one explanation per example, we propose to generate explanations for each possible task output, conditioned on that output. 
Then, we give each answer choice its own input sequence, which includes the task input and an explanation supporting that answer choice.
Finally, a classifier scores each input and output sequence.

Instead of maximizing the likelihood of correct answer tokens, we find that a new learning objective is necessary for training the task model. We renormalize the decoder likelihoods of each answer choice $a_i$ given the encoder input $s_i$. With the set of encoder sequences $S$ and answer choices $A$, we define the probability of each answer choice as:
\vspace{-3pt}
\begin{align*}
p(a_i|A,S) = \frac{p(a_i|s_i)}{\sum_{a_i\in A,s_i\in S}p(a_i|s_i)}
\vspace{-3pt}
\end{align*}
Then we maximize the likelihood of the correct answer choice.

\begin{table}[t]
\small
    \centering
\begin{tabular}{l c l l} 
\toprule
Method & & Task Set & Conditioning \\
\midrule
    T5-Base & & Single-task & -       \\
    \textsc{ST-Re} & & Serial-task  & $e|x$    \\
    \textsc{ST-Ra} & & Serial-task  & $e|x,y$  \\
    \textsc{MT-Re} & & Multi-task   & $e|x$    \\
    \textsc{MT-Ra} & & Multi-task   & $e|x,y$  \\
\bottomrule
 \end{tabular}
 \vspace{-5pt}
\caption{The graphical models and baseline we evaluate. MT and ST refer to multi-task and serial-task, while \textsc{Re} and \textsc{Ra} refer to reasoning and rationalizing.
}
\vspace{-8pt}
\label{table:conditions}
\end{table}

\begin{table*}[t]
\begin{center}
\small
\begin{tabular}{p{0.65\textwidth}p{0.06\textwidth}p{0.045\textwidth}p{0.06\textwidth}p{0.045\textwidth}}
\toprule 
& \multicolumn{2}{c}{Model} & \multicolumn{2}{c}{Human}  \\
\cmidrule(lr){2-3} \cmidrule(lr){4-5}
Input, Output, and Explanation & Leaking? & LAS & Leaking? & LAS \\
\midrule
Question: Marathoners feel fatigued after running twenty six miles, but some that have pushed them self too hard might be prone to what? & \multirow{4}{*}{Yes} & \multirow{4}{*}{1} &  \multirow{4}{*}{Yes} & \multirow{4}{*}{1}\\
Choices: A. passing out; B. death; \textbf{C. exhaustion} \\
\textsc{STRa} explanation: if you are running too hard, you are likely to be exhausted. \\
\midrule
Question: When are people buying products more? & \multirow{4}{*}{No} & \multirow{4}{*}{-1} &  \multirow{4}{*}{No} & \multirow{4}{*}{-1}\\
Choices: \textbf{A. economic boom}; B. disagreements; C. being able to use \\
\textsc{Human} explanation: being able to use. \\
\toprule
\end{tabular}
\end{center}
\vspace{-10pt}
\caption{
Two example data points from CQA with \textsc{Human} or \textsc{STRa} label (\textbf{bold} in text) and explanation. We give leakage indicators and example-level LAS scores from both model-based (T5) and human simulators (see Section \ref{sec:sim_metric}). More examples can be found in Table~\ref{tab:more-exp-examples}.}
\vspace{-12pt}
\label{tab:exp-examples}
\end{table*}

\paragraph{\textsc{MT-Re}.} The alternative to using explanations as task model inputs is to use them as supervision in a multi-task framework. 
As a counterpart to \textsc{ST-Re}, we test a reasoning multi-task model, where explanations are conditioned only on the task input (shown in Figure \ref{fig:t5}).
We use a single task-specific word prepended to the input sequence so that the encoder hidden states will be tailored to either the task or explanation generation. 
For this model, the multi-task learning objective mixes a label prediction loss $\mathcal{L}_{task}$ (for the task itself), and a language modeling loss $\mathcal{L}_{LM}$ (for explanation generation): 
\vspace{-3pt}
\begin{align*}
    \mathcal{L}_{MT}=\alpha\mathcal{L}_{task}+(1-\alpha)\mathcal{L}_{LM},
    \vspace{-3pt}
\end{align*}
where $\alpha$ is the mixing ratio to be tuned on development set. We reach a value of $\alpha = .5$ on both datasets when tuning for task accuracy.

\paragraph{\textsc{MT-Ra}.} Represented in Figure \ref{fig:t5}, \textsc{MT-Ra} is a multi-task model where explanations are conditioned on the model output. This approach originates in \citet{camburu_e-snli:_2018}, where it is introduced as an LSTM-based model. 
As above, we use a task mixing weight of $\alpha = .5$ for both datasets.

\section{LAS: Leakage-Adjusted Simulatability}
\label{sec:sim_metric}

\begin{figure}[t]
\centering
 \includegraphics[width=.33\textwidth]{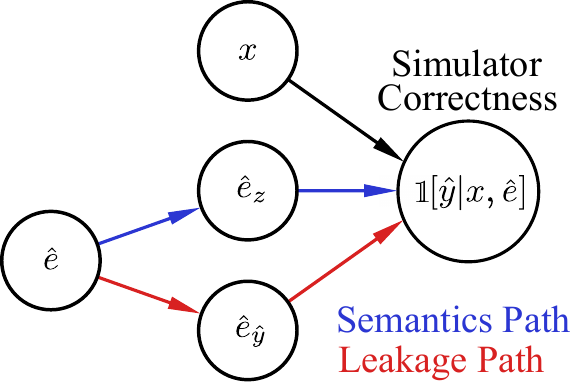}
 \vspace{-3pt}
\caption{
Causal diagram of model simulation. The simulator prediction's correctness, $\mathbbm{1}[ \hat{y}|x,\hat{e}]$, is influenced by three variables: (1) the task model input, (2) the model explanation's semantic content, $\hat{e}_z$, and (3) whether the explanation leaks the model output, $\hat{e}_{\hat{y}}$
} 
\vspace{-7pt}
\label{fig:explanation_effect}
\end{figure}

While many motivations drive humans' explanations for their behavior, we consider one central purpose to be helping others understand one's internal reasoning.
This notion is captured by the concept of simulatability \cite{doshi-velez_towards_2017}. A model is simulatable to the extent 
that an observer, or simulator, can predict its outputs.
The simulator can be either a human or a learned model; we will consider both settings.
From this perspective, one might use the simulator's accuracy to measure explanation quality. With task inputs $X=\{x_i\}$, model outputs $\hat{Y}=\{\hat{y}_i\}$, model explanations $\hat{E}=\{\hat{e}_i\}$, simulator correctness as $\mathbbm{1}[\hat{y}_i|x_i, \hat{e}_i]$,\footnote{For the remainder of the paper, we use the indicator function in this way to describe the correctness of predictions, which is a slight abuse of notation for the sake of brevity.} the accuracy is defined as: 
\vspace{-3pt}
\begin{align*}
\text{Acc}(\hat{y}|x, \hat{e}) = \frac{1}{N}\sum_{i=1}^{N}\mathbbm{1}[\hat{y}_i|x_i, \hat{e}_i]
\vspace{-3pt}
\end{align*}
However, this measure fails to distinguish between different ways in which the simulator can successfully predict the task model output, as shown in the causal diagram in Figure \ref{fig:explanation_effect}. 
We suggest that the simulator's success does not reflect explanation quality when (1) the simulator can guess behavior correctly from the input $x$ alone, or (2) the explanation $\hat{e}$ directly restates the task model output, i.e., leaking the label to the simulator. 
What we are truly looking for in explanations is that they provide semantic content that informs the simulator of the task model's output in the context of its input.
Note that we do not think label leakage means an explanation is bad. Explanations will leak more often than not, as human explanations leak about 85\% of the time for CoS-E and about 97\% of the time for e-SNLI (estimated by T5 simulator). Instead, we think the more important aspect is to evaluate the explanation's semantic content. For examples of leaking and nonleaking explanations, see Table~\ref{tab:exp-examples}.

\begin{figure*}[t]
\centering
 \includegraphics[width=.88\textwidth]{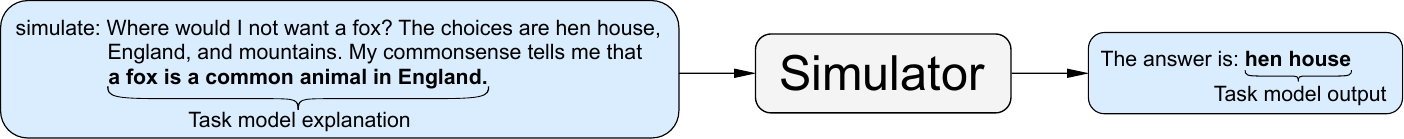}
 \vspace{-4pt}
\caption{A simulator model predicts a task model output, given its input and a model-generated explanation.
} 
\label{fig:simulation}
\vspace{-7pt}
\end{figure*}

To deal with issue (1) above, we introduce an input-only baseline and measure the effect of an explanation on simulatability as 
$\mathbbm{1}[\hat{y}|x,\hat{e}] - \mathbbm{1}[\hat{y}|x]$.
To resolve the issue (2), we propose to control for a label leaking variable, which has the effect of blocking that causal pathway
\cite{pearl_causal_2009}. We do so by using a proxy variable for label leakage, which is an indicator variable for if the simulator can predict $\hat{y}$ solely from $\hat{e}$. The correctness of this prediction suggests that the explanation gives away the answer directly. With this approach, we can estimate explanations' leakage-controlled effect on simulatability by (1) grouping data by the level of explanation label leakage, (2) computing the average difference
$\mathbbm{1}[\hat{y}|x,\hat{e}] - \mathbbm{1}[\hat{y}|x]$
within each leakage group, and (3) taking the raw average of the effect across groups (to avoid favoring the larger subset). 
Note that there are only two levels of label leakage, $\mathbbm{1}[\hat{y}|\hat{e}]=1$ (leaking) and $\mathbbm{1}[\hat{y}|\hat{e}]=0$ (nonleaking), and we use model correctness rather than probabilities since T5 probabilities are uncalibrated. 

Now with simulator correctness as $\mathbbm{1}[\hat{y}_i|x_i, \hat{e}_i]$ or $\mathbbm{1}[\hat{y}_i|x_i]$, and our leakage indicator as $k_i = \mathbbm{1}[\hat{y}_i|\hat{e}_i]$, we write our Leakage-Adjusted Simulatability (LAS) metric as: 
\begin{align*}
\vspace{-3pt}
    \text{LAS}_{0} & = \frac{1}{n_{0}} \sum_{i:k_i=0}\big( \mathbbm{1}[\hat{y}_i|x_i, \hat{e}_i] - \mathbbm{1}[\hat{y}_i|x_i]\big) \\
     \text{LAS}_{1} & = \frac{1}{n_{1}} \sum_{i:k_i=1}\big( \mathbbm{1}[\hat{y}_i|x_i, \hat{e}_i] - \mathbbm{1}[\hat{y}_i|x_i]\big) \\
    \text{LAS} & = \frac{1}{2} (\text{LAS}_{0} + \text{LAS}_{1})
    \vspace{-3pt}
\end{align*}
where $n_{0}$ and $n_{1}$ are the number of examples in nonleaking and leaking groups respectively. 
We use a pretrained T5-Base model as a proxy for a human simulator (depicted in Figure~\ref{fig:simulation}). 
This approach has the advantage of scaling across large datasets with uniform quality in predictions, and, as described in Section \ref{sec:two_agent}, it enables directly optimizing explanations for simulatability. We validate this choice of proxy with two human subject experiments (see Section \ref{sec:human_eval}). 
Simulator models are trained with task model outputs as labels and $x$ and $\hat{e}$ combined into input sequences. In order to make sure the simulator makes good use of both $x$ and $\hat{e}$, we randomly dropout either $x$ or $\hat{e}$ from the input during training. When testing, the simulator's correctness on each example is $\mathbbm{1}[\hat{y}_i|x_i, \hat{e}_i]$, and we obtain $\mathbbm{1}[\hat{y}_i|x_i]$ and $\mathbbm{1}[\hat{y}_i|\hat{e}_i]$ by dropping $\hat{e}_i$ or $x_i$ from the input.

We will compare LAS and $\text{Acc}(\hat{y}|x, \hat{e})$ for explanations from the models introduced above as well as human explanations. We discuss the relationship with human experiments for both metrics in Section~\ref{sec:human_eval}. In analysis to follow, we will also refer to example-level LAS scores, which are given as $\mathbbm{1}[\hat{y}|x,\hat{e}] - \mathbbm{1}[\hat{y}|x]$ and take values -1, 0, or 1 (see Table \ref{tab:exp-examples} for examples). 
Lastly, while we use a binary proxy for label leakage, a continuous measure can be obtained from $p(\hat{y}|\hat{e})$. After calibrating the simulator probabilities via Platt scaling \cite{platt}, we perform a sensitivity analysis of our results for bin counts between 2 and 100: LAS estimates typically vary by less than 1 point across bin counts. For further details, see Appendix~\ref{appendix:continuous-las}.

\section{Multi-Agent Explanation Optimization}

\label{sec:two_agent}
In this section, we explore an approach to optimizing explanations for LAS, rather than just relying on a standard language modeling loss to produce explanations. The approach is naturally framed as a multi-agent game.
Note that we do not aim to improve model accuracy or explanations' BLEU scores in these experiments.

\begin{table*}[th]
    \centering
    \small
\begin{tabular*}{\textwidth}{l @{\extracolsep{\fill}} l rrr rrr} 
\toprule
& & \multicolumn{3}{c}{SNLI} & \multicolumn{3}{c}{\textsc{CQA}}  \\
\cmidrule(lr){3-5} \cmidrule(lr){6-8}
Explanations & & LAS Score (CI) & Acc($\hat{y} \ | \ x,\hat{e}$) & BLEU & LAS Score (CI) & Acc($\hat{y} \ | \ x,\hat{e}$) & BLEU \\ 
\midrule
    \textsc{Human} & & 4.31 (1.97) & 98.36 & - & 14.73 (3.84) & 90.11 & - \\
    \textsc{MT-Re} & & -15.83 (1.81) & 93.72 & 19.54 & -7.07 (3.59) & 81.05 & 6.33 \\
    \textsc{MT-Ra} & & 4.34 (4.12) & 99.97 & 19.41 & -1.31 (4.04) & 92.31 & 5.43 \\
    \textsc{ST-Re} & & 0.55 (0.87) & 93.87 & 19.96 & 3.76 (1.83) & 82.21 & 7.12 \\
    \textsc{ST-Ra} & & 6.74 (4.53) & 99.84 & 20.94 & 10.32 (3.39) & 88.53 & 7.14 \\
    \addlinespace[1pt]
    \textsc{Multi-Agent} \\
    \quad \textsc{MT-Re} (SGD) & & -10.08 (1.72) & 94.14 & 16.74 & -6.32 (3.27) & 76.63 & 4.44 \\
    \quad \textsc{MT-Ra} (SGD) & & 3.03 (4.72) & 99.89 & 16.61 & 3.08 (3.79) & 87.68 & 4.43 \\
    \quad \textsc{MT-Re} (RL) & & -10.80 (1.51) & 93.45 & 15.41 & -5.04 (3.55) & 84.00 & 2.15 \\
    \quad \textsc{MT-Ra} (RL) & & -0.61 (0.45) & 93.05 & 9.83 & -9.15 (2.95) & 77.47 & 3.54 \\
\bottomrule
 \end{tabular*}
 \vspace{-5pt}
\caption{Evaluations of human and model-generated explanations by LAS score, overall simulator accuracy, and BLEU. 95\% confidence intervals as calculated by bootstrap are shown in parentheses \cite{efron1994introduction}.}
\label{table:sim_results}
\vspace{-9pt}
\end{table*}

In our game, there are two agents. The first is a \textbf{task model} that predicts labels and generates explanations jointly. Here, we use \textsc{MT-Re} or \textsc{MT-Ra}. The second agent is a \textbf{simulator} model that predicts the task model's output $\hat{y}_i$ given its explanation $\hat{e}_i$ and the model input $x_i$, matching the previous simulation format shown in Figure~\ref{fig:simulation}. These two agents are jointly trained during the multi-agent training procedure. The objective of the simulator is the same as discussed in the above section, which is to predict $\hat{y}_i$ given $x_i$ and $\hat{e}_i$, and we randomly dropout $x_i$ or $\hat{e}_i$ to ensure they are both being used. 
As in Section~\ref{sec:methods_modeling_explanations}, the task model learns to perform the task (minimizing $\mathcal{L}_{task}$) and generate explanations (minimizing $\mathcal{L}_{LM}$) via supervision from ground-truth labels and human explanations. 
Here, the task model also tries to minimize the simulator's loss through its explanations.
The chief computational challenge with this approach is that explanations are sampled by greedy decoding, and thus the loss is not differentiable with respect to the task model. 
We explore two optimization methods circumventing this issue: Approximate SGD via argmax relaxation \cite{maddison_concrete_2016} and REINFORCE \cite{williams_simple_1992}. 
Our aim is for explanations to better communicate the task model's reasoning process, without adopting the trivial solution, i.e., directly stating its output. Thus while we optimize explanations for simulatability, we also penalize label leakage, which we formalize below. 
Note that the task model's predictions are not optimized to agree with the simulator; only its explanations are optimized. 

\vspace{3pt}
\noindent\textbf{Approximate SGD.}
With a simulator model $p_\phi$, the simulatability loss term for explanations is
\begin{align*}
\mathcal{L}_{exp} = & -\frac{1}{N} \sum_{i=1}^{N} \big(\alpha \log p_\phi(\hat{y}_i|x_i, \hat{e}_i) \\  
    & \hspace{1.2cm} - (1-\alpha)\log p_\phi(\hat{y}_i|\hat{e}_i) \big)
\end{align*}
where $\alpha$ is a mixing weight between terms. To differentiate through the greedy decoding for explanation sampling, we use one half of the Gumbel-Softmax trick \cite{maddison_concrete_2016}. During the forward pass in training, the argmax is used as normal, while during the backward pass, we relax the argmax to a softmax with temperature 1 for purposes of computing gradients. 

\vspace{3pt}
\noindent\textbf{Reinforce.} Our second approach is to use the REINFORCE  RL algorithm proposed by \citet{williams_simple_1992}. Here we take the simulator's output probabilities as a reward for the task model. Now with the same goals as above, we define the reward for 
$x_i$ as $r_i = \alpha p_\phi(\hat{y}_i|x_i, \hat{e}_i) - (1-\alpha)p_\phi(\hat{y}_i|\hat{e}_i)$.
Then, the $\mathcal{L}_{exp}$ for task model $p_{\theta}$ is defined as:
\vspace{-2pt}
\begin{align*}
   \mathcal{L}_{exp} &=  \frac{1}{N}\sum_{i=1}^{N} - r_i \log p_\theta(\hat{e_i} | x_i, \hat{y}_i)
\end{align*}
\vspace{-2pt}Finally, with either method, the full learning objective of the task model is $\mathcal{L}_{TaskModel} =\lambda_1\mathcal{L}_{task} + \lambda_2\mathcal{L}_{LM} + \lambda_3 \mathcal{L}_{exp}$.
The tuning procedure and values for mixing weights are given in Appendix~\ref{sec:hyperparamter_tuning}.

%% file: experiments_and_results.tex
\section{Experimental Results}
\label{sec:experiments}

Here, we discuss experiments conducted with each method using two (English) datasets:
The first is the \textsc{CommonSenseQA} (CQA) dataset of \citet{talmor_commonsenseqa_2019}, with explanations collected by \citet{rajani_explain_2019} to make a combined CoS-E dataset (examples in Table \ref{tab:exp-examples}). We use the Version 1.0 of this dataset, since it has higher quality explanations than Version 1.1.\footnote{In Version 1.1, about 20\% of explanations belong to a small set of duplicates unrelated to the data point. See \url{https://github.com/salesforce/cos-e/issues/2}.} CQA has approximately 8k/1k/1k train/dev/test data points, while NLI has roughly 549k/10k/10k train/dev/test points. 
Note that, in the main paper, we report results using 10\% of the SNLI training data, due to computational demands of tuning multi-task models (1 week for convergence with 100\% data), and we report CQA dev results since human explanations are not available for test data. See Tables \ref{table:nli_full} and \ref{table:acc_alternative_results} in the Appendix for results for CQA test data and SNLI with full training data, where we confirm the results discussed here. For the model selection procedure and further training details, see Appendix~\ref{appendix:model_selection}, and for robustness checks of LAS scores across seeds and simulator architectures, see Appendix~\ref{appendix:robustness_seed_model}.

\subsection{Automatic Explanation Evaluation}

Below we describe key conclusions from our evaluation of leakage-adjusted simulatability (LAS), and we show results alongside overall simulator accuracy $\text{Acc}(\hat{y}|x, \hat{e})$ and BLEU in Table \ref{table:sim_results}.

\vspace{5pt}
\noindent \textbf{Humans vs. Models.} Some models do achieve roughly human-level LAS scores for CQA and NLI. First, we find that human explanations are helpful to models: we estimate that explanations improve humans' simulatability by 4.31 percentage points for SNLI and by 14.73 points for CQA. Our \textsc{ST-Ra} method performs similarly to humans on both datasets. On SNLI, \textsc{MT-Ra} also achieves about human performance. We emphasize that this does not mean these models match human explanations in every respect. Rather, the semantics of the explanations have a similar effect on simulator accuracy as human explanations in our experimental settings. Additionally, we note that scores across datasets are not directly comparable since they depend on the underlying difficulty of the task.

\vspace{5pt}
\noindent\textbf{\textsc{Re} vs. \textsc{Ra}.}
Rationalizing models outperform their reasoning counterparts on both datasets. For \textsc{MT-Re}, the drop in LAS stems from non-leaking explanations -- these explanations tend to mislead the simulator, meaning $p(\hat{y}|x,\hat{e})$ is inaccurate.
For \textsc{ST-Re}, explanations tend to leak for examples where it is already easy to guess model behavior from $x$, i.e. $p(\hat{y}|x)$ sets a high baseline.

\vspace{5pt}
\noindent\textbf{BLEU vs. Simulatability.} BLEU is not correlated with our LAS metric, which supports our conjecture that BLEU does not reflect the effect of explanations on simulatability. LAS also does not correlate with the simulator accuracy, Acc($\hat{y} | x,\hat{e}$), which is expected given how the simulator accuracy is heavily influenced by explanation leakage.

\subsection{Human Validation of LAS}
\label{sec:human_eval}

We validate our model proxy variables with two human evaluations, an expert simulation experiment, and a crowdsourced subjective rating test.  

\begin{table}[t]
    \centering
    \small
\begin{tabular}{@{}c@{}}
\toprule
\begin{tabular}{c @{\extracolsep{\fill}} l c c} 
\hspace{-8pt} \textbf{Leakage} & & \multicolumn{2}{c}{Human} \\
\cmidrule(lr){3-4}
 Model & & 0 & 1 \\ 
\midrule
    0 & & 127 & 87 \\
    1  & & 45 & 341 \\
    && & \\
 \end{tabular}
\quad \quad
\begin{tabular}{c @{\extracolsep{\fill}} l c c c} 
\hspace{-17pt}\textbf{LAS} & & \multicolumn{3}{c}{Human} \\
\cmidrule(lr){3-5}
 Model & & -1 & 0 & 1 \\ 
\midrule
    -1 & & 23 &  56 &  6 \\
    0  & & 29 & 278 & 49 \\
    1 & &  5 & 104 & 50  \\
 \end{tabular}
\tabularnewline
\bottomrule
\end{tabular}
\caption{Correlation between model-based and human variables resulting from the expert simulation analysis. For the leakage variable, Spearman's rank correlation is $\rho = 0.53 \ (p<1\mathrm{e}{-15})$. For the example-level LAS, the rank correlation is $\rho = 0.29 \ (p<1\mathrm{e}{-12}$). }
\label{table:sim_validation}
\vspace{-8pt}
\end{table}

\vspace{3pt}
\noindent\textbf{Expert Simulation.}
We (meaning the first three authors as expert annotators) validate our use of models as simulators of both model-generated and human explanations by manually playing the role of the simulator for 600 data points. With effectively the same design as our automatic metric computation, we simulate humans and our \textsc{ST-Ra} model with both datasets, only with no training period in this case. Each annotator is randomly assigned a role for each data point (whether they see the input, explanation, or both), and points are sampled such that an annotator never sees the same point in different roles. The sample is roughly balanced across the strata of our model's proxy variables. We note that ideally, we would use only expert human simulators instead of proxies, though even annotating less than 1\% of the data across conditions required 1800 individual responses. 

The correlations between proxy variables and our own are shown in Table~\ref{table:sim_validation}. We group the data across subsets (e.g., explanation source and dataset) since the trends were similar between them.
We find a strong correlation between the leakage proxy variable and the human leakage variable, with a Spearman rank correlation of $\rho = 0.53 \ (p<1\mathrm{e}{-15})$, and we observe a moderate correlation between the model-based and human example-level LAS, $\rho = 0.29 \ (p<1\mathrm{e}{-12})$ \cite{cohen1988}. 

The disagreements are concentrated in false negatives for leakage, where we identify leaking explanations when the model does not. With LAS, model scores of -$1$ and $1$ often end up as a human 0, meaning that an explanation confuses the model but not the human rater (for -1), or the human can predict based on the input alone when the model cannot (for 1). 
Because of this tendency toward 0, human LAS will shrink slightly toward 0 in expectation, relative to the model LAS (see row-normalized Table \ref{table:sim_validation_norm} in Appendix).
We also observe a degree of \emph{pragmatic drift} between models and humans. \citet{lazaridou2020} operationalize this as the difference in performance between human and model listeners in a reference game. Similarly, we can use simulator accuracy given the input and explanations. We find that humans are better simulators of humans, and models are better at predicting model outputs. Across datasets and simulators, the difference in accuracies is 12.83 percentage points on average. 

Lastly, one may notice from Table \ref{table:sim_validation} that our predictions of the human label are sometimes wrong. In fact, our own task accuracy is 70\% ($\pm 7.33$) for SNLI and 72\% for CQA ($\pm 7.19$). These accuracies are similar to those obtained by \citet{pavlick2019} when re-annotating the SNLI dataset. Interestingly, they find that tasks such as these can have distributions over labels under human annotation, rather than consensus. 

\vspace{3pt}
\noindent\textbf{Human Subjective Quality Ratings.}
We collect human ratings from Mechanical Turkers for 200 test examples for both CQA and SNLI. Each example includes shuffled, unlabeled explanations (one from each model, plus humans, for a total of five), which we ask workers to separately rate on a 5-point Likert scale. After collecting 3 responses per item, we apply a worker quality filter, obtaining 902 ratings total. Further collection details are provided in Appendix~\ref{appendix:mturk}.

We investigate whether LAS and simulator accuracy are correlated with human explanation ratings. For each example, we obtain human ratings, the example's LAS score $\mathbbm{1}[\hat{y}|x, \hat{e}] - \mathbbm{1}[\hat{y}_i|x_i]$ (taking values -1,0,1), and simulator prediction accuracies, $\mathbbm{1}[\hat{y}|x, \hat{e}]$, $\mathbbm{1}[\hat{y}|x]$, and $\mathbbm{1}[\hat{y}|\hat{e}]$ (taking values 0 or 1). 

Human rating trends across example-level LAS scores are shown in Tables \ref{table:human_las}. A first observation is that LAS scores do not correlate well with human ratings.
Curiously, though, simulator accuracies correlate with human ratings. We show these trends in Table \ref{table:human_sim}, along with regression coefficients for predicting ratings from simulator accuracies. For both datasets, $\mathbbm{1}[\hat{y}|\hat{e}]$ best correlates with human ratings and the association with 
$\mathbbm{1}[\hat{y}|x, \hat{e}]$
is only significant for SNLI. Since good explanations tend to leak the label, it is not surprising that ratings correlate with label leakage. However, it is surprising that this association is stronger than the relationship with overall accuracy, $\mathbbm{1}[\hat{y}|x, \hat{e}]$. Together, these results help explain why models may struggle to learn from human explanations, since models may focus on label leakage in human explanations at the expense of other information. They may also suggest that to collect human ratings that do not correlate with label leakage, a highly controlled environment for human ratings may be required.

\begin{table}[t]
    \centering
    \small
\begin{tabular}{l c c c} 
\toprule
& \multicolumn{3}{c}{Example-Level LAS Score} \\
\cmidrule(){2-4}
Data \& Leakage & -1 & 0 & 1 \\ 
\midrule
\hspace{-2mm}CQA: Leaking & 2.39 (.36) & 2.65 (.08) & 2.58 (.15)  \\
\hspace{-2.05mm}\hspace{7.5mm} Non-leaking & 2.31 (.21) & 2.40 (.10) & 2.28 (.34) \\
\addlinespace[3pt]
\hspace{-2mm}SNLI: Leaking & 2.96 (.45) & 3.25 (.06) & 3.18 (.15)  \\
\hspace{-2mm}\hspace{7.8mm} Non-leaking & 2.78 (.31) & 2.94 (.12) & 2.61 (.46) \\
\bottomrule
 \end{tabular}
 \vspace{-5pt}
\caption{
Human explanation ratings grouped by dataset, label leakage. 95\% confidence intervals in parentheses.
}
\label{table:human_las}
\vspace{-9pt}
\end{table}

\subsection{Accuracy-Interpretability Trade-off}
Past works on model interpretability have observed trade-offs between accuracy and model constraints for interpretation purposes \cite{bastings-2019, jain2020}. Yet, \citet{rudin_stop_2019} and \citet{jacovi2020aligning} argue that we need not always face such a trade-off. Our findings provide quantitative evidence supporting these prior qualitative arguments. We observe consistently small changes in accuracy for our four models, and the largest changes, -.47 ($p=.3124$) for SNLI and -2.10 for CQA ($p=.3272$), are not statistically significant. 
We also test methods using human explanations purely for improving accuracy, e.g., through Masked Language Modeling objectives that have been successful for pretraining models. We find that this objective does not lead to statistically significant accuracy improvements, suggesting models still struggle to truly learn from human explanations (results are shown in Table \ref{table:acc_alternative_results}).

\begin{table}[t]
    \centering
    \small
\begin{tabular}{l c c c c} 
\toprule
& \multicolumn{2}{c}{Simulator Correctness} & \multicolumn{2}{c}{Regression Coef.} \\
\cmidrule(lr){2-3}\cmidrule(lr){4-5}
Prediction & 0 & 1 & $\beta$ & $p$ \\ 
\midrule
\hspace{-3.2mm} CQA: $\hat{y}|x,\hat{e}$ & 2.34 (.11) & 2.60 (.06)   & .14 & .07 \\
\hspace{5.15mm} $\hat{y}|x$ & 2.38 (.09) & 2.63 (.07)  & .09 & .20 \\
\hspace{5.15mm} $\hat{y}|e$ & 2.44 (.10) & 2.58 (.07)  & .21 & $<$.001 \\
\addlinespace[3pt]
\hspace{-3.2mm} SNLI: $\hat{y}|x,\hat{e}$ & 2.85 (.14) & 3.22 (.05)  & .20 & .03  \\
\hspace{5.5mm} $\hat{y}|x$ & 2.90 (.11) & 3.24 (.06)  & .10 & .15 \\
\hspace{5.5mm} $\hat{y}|e$ & 3.02 (.11) & 3.21 (.08)  & .27 & $<$.001 \\
\bottomrule
 \end{tabular}
 \vspace{-4pt}
\caption{Human ratings broken down by dataset and simulator prediction, shown alongside regression results.
95\% confidence intervals in parentheses.}
\label{table:human_sim} 
\vspace{-8pt}
\end{table}

\subsection{Multi-Agent Game}

Multi-agent game results appear in Table \ref{table:sim_results}, though we note that RL results should be cautiously interpreted as we observe unstable training behavior from this method. We find that optimization with SGD can reduce label leakage (from, e.g., 85.58\% to 75.21\% for CQA \textsc{MT-Ra}) while slightly improving LAS scores, but only one of four changes in LAS scores is statistically significant, for \textsc{MT-Re} on SNLI. This approach does pull BLEU scores down. No statistically significant differences in accuracy are found; the largest change, a 3.37 point drop on CQA, has a $p$-value of $.1287$. We note that this kind of optimization may have the effect of increasing pragmatic drift, as is found for jointly optimized agents in \cite{lazaridou2020}.

%% file: appendix.tex
\section{Experimental Details}
\label{appendix:experimental_details}

\subsection{Datasets and Examples}
\label{sec:datasets}

We conduct experiments with each method using two datasets. The first is the \textsc{CommonSenseQA}\footnote{https://www.tau-nlp.org/commonsenseqa} dataset of \citet{talmor_commonsenseqa_2019}, with explanations collected by \citet{rajani_explain_2019} to make a combined CoS-E dataset.\footnote{https://github.com/nazneenrajani/CoS-E}
We opt for the Version 1.0 of this dataset since it has higher-quality explanations than Version 1.1.\footnote{In Version 1.1, 20\% of explanations were found to belong to a small set of duplicates that are unrelated to the data point. See \url{https://github.com/salesforce/cos-e/issues/2}.} The dataset split sizes are 7610, 950, and 940 for the train, dev, and test, respectively. Next, we use the e-SNLI dataset of \citet{camburu_e-snli:_2018},\footnote{https://github.com/OanaMariaCamburu/e-SNLI} which includes explanations for the SNLI benchmark \cite{bowman_large_2015}.\footnote{https://nlp.stanford.edu/projects/snli/} The split sizes are 549,339, 9842, and 9824, for train, dev, and test. Three explanations per data point are available for the test data in e-SNLI; to compute BLEU, we use the first explanation in the data for each data point; we use the \texttt{sacrebleu} Python package \cite{post2018call}.\footnote{https://github.com/mjpost/sacreBLEU}

Note that explanations for the CQA test split were not collected for the CoS-E dataset, as the CQA test split itself is withheld as a leaderboard test set. Meanwhile, we report results using 10\% of the SNLI training data, since training our multi-task T5 models with the full e-SNLI dataset can take over 24 hours per epoch on a single T4 GPU. These accuracy results are shown in Table \ref{table:acc_results}. We report test set statistics here for simulation-related experiments for CQA, shown in Table \ref{table:sim_results}, along with dev statistics for SNLI. Trends across models remain the same as with the data split statistics reported in the main paper. In Table \ref{table:nli_full}, we confirm trends observed with the SNLI training data subset using models trained with the entire dataset. Finally, Table~\ref{tab:more-exp-examples} shows additional examples from CQA and SNLI plus model-generated explanations.

\begin{table*}[t]
\begin{center}
\small
\begin{tabular}{p{0.65\textwidth}p{0.06\textwidth}p{0.045\textwidth}p{0.06\textwidth}p{0.045\textwidth}}
\toprule 
& \multicolumn{2}{c}{Model} & \multicolumn{2}{c}{Human}  \\
\cmidrule(lr){2-3} \cmidrule(lr){4-5}
Input, Output, and Explanation & Leaking? & LAS & Leaking? & LAS \\
\midrule
Question: Marathoners feel fatigued after running twenty six miles, but some that have pushed them self too hard might be prone to what? & \multirow{4}{*}{Yes} & \multirow{4}{*}{1} &  \multirow{4}{*}{Yes} & \multirow{4}{*}{1}\\
Choices: A. passing out; B. death; \textbf{C. exhaustion} \\
\textsc{STRa} explanation: if you are running too hard, you are likely to be exhausted. \\
\midrule
Question: Where is likely to not just have a kosher restaurant? & \multirow{4}{*}{Yes} & \multirow{4}{*}{0} &  \multirow{4}{*}{No} & \multirow{4}{*}{0}\\
Choices: \textbf{A. new york city}; B. jewish neighborhoods; C. jerusalem \\
\textsc{Human} explanation: kosher restaurant is not in new york city. \\
\midrule
Question: When are people buying products more? & \multirow{4}{*}{No} & \multirow{4}{*}{-1} &  \multirow{4}{*}{No} & \multirow{4}{*}{-1}\\
Choices: \textbf{A. economic boom}; B. disagreements; C. being able to use \\
\textsc{Human} explanation: being able to use. \\
\midrule
Question: John bought a new water hose.  But he found his old one near his car.  Where did he find the old one? & \multirow{4}{*}{Yes} & \multirow{4}{*}{1} &  \multirow{4}{*}{Yes} & \multirow{4}{*}{0}\\
Choices: A. garden shed; B. hardware store; \textbf{C. garage} \\
\textsc{STRa} explanation: garage is the only place where you can find old water hoses. \\
\midrule
\midrule
Premise: A man of the cloth puts a black substance on a man 's forehead. & \multirow{4}{*}{Yes} & \multirow{4}{*}{1} &\multirow{4}{*}{Yes} & \multirow{4}{*}{1}\\
Hypothesis: The men are at church. \\
Choices: A. entailment; \textbf{B. neutral}; C. contradiction \\
\textsc{Human} explanation: You can not infer they are at church . \\
\midrule
Premise: One tan girl with a wool hat is running and leaning over an object , while another person in a wool hat is sitting on the ground. & \multirow{4}{*}{Yes} & \multirow{4}{*}{0} &\multirow{4}{*}{Yes} & \multirow{4}{*}{0}\\
Hypothesis: A boy runs into a wall. \\
Choices: A. entailment; B. neutral; \textbf{C. contradiction} \\
\textsc{STRa} explanation: A girl is not a boy. \\
\midrule
Premise: A man dressed in a light blue shirt dumping items from a bin into another bin , while standing in a room full of food donations. & \multirow{4}{*}{Yes} & \multirow{4}{*}{-1} &\multirow{4}{*}{Yes} & \multirow{4}{*}{-1}\\
Hypothesis: Foods are not stored in room by a man. \\
Choices: A. entailment; B. neutral; \textbf{C. contradiction} \\
\textsc{STRa} explanation: Food donations are not stored. \\
\midrule
Premise: Taking a break to watch some TV & \multirow{4}{*}{No} & \multirow{4}{*}{-1} &\multirow{4}{*}{No} & \multirow{4}{*}{0}\\
Hypothesis: Taking a neverending break \\
Choices: A. entailment; B. neutral; \textbf{C. contradiction} \\
\textsc{Human} explanation: Some TV is not enough to be on a neverending break. \\
\toprule
\end{tabular}
\end{center}
\vspace{-10pt}
\caption{
Example data points from both CQA and SNLI with \textsc{Human} or \textsc{STRa} label (\textbf{bold} in text) and explanation. Leakage predictions and example-level LAS scores from both model-based (T5) and human simulators are given. }
\vspace{-12pt}
\label{tab:more-exp-examples}
\end{table*}

\subsection{Hypothesis Testing}

We describe results as statistically significant when $p$-values are below $.05$, where $p$-values are calculated by bootstrap for LAS, a difference in the binomial means test for model accuracies, and by linear regression with i.i.d. normal noise for associations between human ratings and simulator correctness. Note that confidence intervals for LAS vary in width based on how many data points are in each leakage bin. 
With the expert evaluation, we compute Spearman's rank correlation between proxy and human simulation variables (with a corresponding $p$-value). For our data, the results are nearly identical to Pearson's linear correlation and Kendall's Tau. 

\subsection{Model Selection and Training Details}
\label{appendix:model_selection}

Our model selection procedure is to train each task model five times with differing seeds, then select the model with the best development performance. We train one simulator model per condition. Since the two-agent experiments have far increased computational load, we run one seed using a T5-Small during training, selecting the best task model according to its LAS with this weaker simulator. Afterward, we retrain with a T5-Base simulator. 

Our training procedures result in the following (approximate) experimental times for each model when training on a single NVIDIA T4 GPU. With a T5-Base model and CQA data, our baseline takes about 10 hours for 20 epochs; \textsc{ST-Re} about 10 hours for 20 epochs; \textsc{ST-Ra} about 20 hours for 20 epochs; \textsc{MT-Re} about 12 hours for 20 epochs; \textsc{MT-Ra} about 12 hours for 20 epochs. Multi-agent RL optimization with a T5-Small simulator takes about 16 hours for 10 epochs, and SGD takes 24 hours for 10 epochs. Now with a T5-Base model and SNLI data (using 10\% of the training data), our baseline takes about 24 hours for 10 epochs; \textsc{ST-Re} about 24 hours for 10 epochs; \textsc{ST-Ra} about 48 hours for 10 epochs; \textsc{MT-Re} about 30 hours for 10 epochs; \textsc{MT-Ra} about 30 hours for 10 epochs. Multi-agent RL optimization with a T5-Small simulator takes about 3 days for 5 epochs, and SGD takes 5 days for 5 epochs. Using the full SNLI dataset, the baseline took four days to train five epochs, and either MT model took 5 days for 5 epochs. We train generators for the ST conditions for 5 epochs on the 10\% subset, which takes under 6 hours. Note that to follow our model selection procedure, experimental times should be multiplied by five here, and further extended to include training simulators.

Lastly, we note that T5-Base has 220 million parameters, while T5-Small as 60 million parameters \cite{raffel_exploring_2019}. In general, this means our model sizes are 220 million parameters, although, for multi-agent training, our effective model size is 280 million parameters. 

\subsection{Training Simulator Models}

When training simulators, it is critical that the model can approximate the three distributions used in LAS computation: $p_{\phi}(\hat{y}_i|x_i, \hat{e}_i)$, $p_{\phi}(\hat{y}_i| x_i)$, and $p_{\phi}(\hat{y}_i| \hat{e}_i)$. This is achieved by applying dropout at the input token level to either (1) the entire $x$ subsequence, or (2) the entire $\hat{e}$ subsequence. The same proportion of inputs in each batch are affected by the dropout, with the subset being chosen randomly. Without this technique, simulator models rely too heavily on explanations, and when conditioned only on $x$, they underperform baseline models that are trained only with $x$. 
In our multi-agent experiments, we take a nearly identical approach, but we make use of the fact that each of the three simulator predictions is made for each batch ($p_{\phi}(\hat{y}_i|x_i$, $\hat{e}_i)$, $p_{\phi}(\hat{y}_i| x_i)$, and $p_{\phi}(\hat{y}_i| \hat{e}_i)$). That is, we weight these terms in the simulator objective by ratios implied by our dropout technique, rather than using dropout directly. See the Section \ref{sec:hyperparamter_tuning} for the relevant hyperparameters.

\subsection{Hyperparameter Tuning}
\label{sec:hyperparamter_tuning}

For baselines, we tune hyperparameters such as the learning rate and batch size for accuracy, selecting from $[1e-5, 1e-4, 1e-3]$ for LR and $[4,6,12,24,36]$ for batch size, finally using $1e-4$, with CQA batch size $12$ and SNLI batch size $36$.

For multi-task models, we tune the mixing weight $\alpha$ based on task performance, searching over values in $[.3,.4,.5,.6,.7,.8]$, settling on $.5$. 

For simulator models, we tune mixing weights (or dropout proportions) by selecting based on each of the three predictions' accuracies, relative to baseline models trained on one input type only. Specifically, we select based on the max accuracy of the subsequence ($x$ and $e$) predictions (with accuracies added together), under the constraint that models must achieve within 1 percentage point accuracy of the overall $p_{\phi}(\hat{y}_i|x_i, \hat{e}_i)$ accuracy. Now taking $\lambda_{x,e}, \lambda_x,$ and $\lambda_e$ as loss function weights for predictions conditioned on their subscripts, the effective loss function weights for CoS-E data are: $\lambda_{x,e} = .5, \lambda_x = .5$, and $\lambda_e = 0$; and for NLI, we use $\lambda_{x,e} = .4, \lambda_x = .4, \lambda_e = .2$. 

The most complex set-up for tuning is our multi-agent method. Here, we must tune mixing weights for the task, LM, and explanation objectives, as well as the weight for penalizing leaking explanations. First, we tune the task, LM, and simulatability weights directly for overall simulator accuracy, without applying a penalty for leaking. We search each parameter over the range $[.2,.5]$ spaced by .05, with constraints that the three terms must add to 1, task weight must be as high as LM weight, and sim weight must be as high as task weight). Lastly, we tune the $\alpha$ trading off between explanation rewards and penalties by selecting directly for LAS scores; we search the unit interval spaces by .1. For SGD, $\alpha$ is set to $.8$ for CQA and $.9$ for SNLI; the task loss is $.35$, LM loss is $.15$, explanation loss is $.5$, and the simulator model objective adopts the same weights as described above. For RL, this mixing weight $\alpha$ is set to $.8$ for both datasets; the task loss is $.025$, LM loss is $.025$, explanation loss is $.95$, and the simulator model objective also adopts the same weights as described above.

\begin{table}[t]
    \centering
    \small
\begin{tabular}{l c c c} 
\toprule
& \multicolumn{2}{c}{SNLI} & \multicolumn{1}{c}{\textsc{CQA}}  \\
\cmidrule(lr){2-3} \cmidrule(lr){4-4}
\hspace{-2mm}Method & Dev Acc & Test Acc & Dev Acc\\ 
\midrule
    \hspace{-2mm}\textsc{T5-Base} & 88.58 & 88.14 (.63) & 68.84 (2.95)  \\
    \hspace{-2mm}\textsc{MT-Re} & 88.91 & 88.44 (.62) & 69.26 (2.93) \\
    \hspace{-2mm}\textsc{MT-Ra} & 88.95 & 87.98 (.63) & 68.95 (2.94) \\
    \hspace{-2mm}\textsc{ST-Re} & 87.67 & 87.67 (.64) & 66.74 (3.00) \\
    \hspace{-2mm}\textsc{ST-Ra} & 87.69 & 87.69 (.64) & 68.84 (2.95) \\
    \addlinespace[2pt]
    \hspace{-2mm}\textsc{Multi-Agent}\\
    \textsc{MT-Re} (SGD) & 88.24 & 87.94 (.64) & 68.00 (2.97) \\
    \textsc{MT-Ra} (SGD) & 88.04 & 87.68 (.64) & 65.58 (3.02) \\
    \textsc{MT-Re} (RL) & 88.31 & 87.91 (.64) & 68.31 (2.96) \\
    \textsc{MT-Ra} (RL) & 87.99 & 87.72 (.65) & 67.47 (2.98) \\
\bottomrule
 \end{tabular}
\caption{Model accuracies for the CQA and SNLI tasks. Generative models perform as well as non-generative baselines. CQA results are for dev data and SNLI are dfor test.}
\vspace{-6pt}
\label{table:acc_results}
\end{table}

\begin{table*}[th]
    \centering
    \small
\begin{tabular*}{\textwidth}{l @{\extracolsep{\fill}} l rrr rrr} 
\toprule
& & \multicolumn{3}{c}{Dev. SNLI} & \multicolumn{3}{c}{Test \textsc{CQA}}  \\
\cmidrule(lr){3-5} \cmidrule(lr){6-8}
Explanations & & LAS Score (CI) & Acc($\hat{y} \ | \ x,\hat{e}$) & BLEU & LAS Score (CI) & Acc($\hat{y} \ | \ x,\hat{e}$) & BLEU \\ 
\midrule
    \textsc{Human} & & 4.36 (2.10) & 98.40 & - & - & - & - \\
    \textsc{MT-Re} & & -14.08 (1.78) & 94.05 & - & -5.40 (3.73) & 80.00 & - \\
    \textsc{MT-Ra} & & 2.70 (8.59) & 99.92 & - & 2.25 (4.60) & 91.91 & - \\
    \textsc{ST-Re} & & 1.52 (0.90) & 94.44 & - & 2.78 (2.10) & 82.23 & - \\
    \textsc{ST-Ra} & & 7.26 (3.20) & 99.90 & - & 10.33 (3.34) & 86.70 & - \\
    \addlinespace[1pt]
    \textsc{Multi-Agent} \\
    \quad \textsc{MT-Re} (SGD) & & -9.56 (1.64) & 94.44 & - & -2.16 (3.56) & 77.23 & - \\ 
    \quad \textsc{MT-Ra} (SGD) & & 5.06 (5.97) & 99.90 & - & 4.53 (3.51) & 84.79 & - \\ 
    \quad \textsc{MT-Re} (RL) & &  -12.08 (1.51) & 93.52 & - & -6.55 (3.38) & 80.95 & - \\ 
    \quad \textsc{MT-Ra} (RL) & &  -0.52 (0.45) & 93.18  & - & -9.59 (2.93) & 70.31 & - \\ 
\bottomrule
 \end{tabular*}
 \vspace{-5pt}
\caption{Evaluations of human and model-generated explanations by LAS score, overall simulator accuracy, and BLEU. We show the opposite data split relative to the main paper, for reproducibility. 95\% confidence intervals as calculated by bootstrap are shown in parentheses. Confidence intervals are wider when the nonleaking subset is very small, and smaller when leaking and nonleaking subsets are both large.}
\label{table:sim_results_appendix}
\vspace{-3pt}
\end{table*}

\begin{table}[t]
    \centering
    \small
\begin{tabular}{l r r r} 
\toprule
& \multicolumn{3}{c}{Seed} \\
\cmidrule(lr){2-4}
\hspace{-2mm}Method & Seed 1 & Seed 2 & Seed 3\\ 
\midrule
    SNLI\\
    \hspace{2mm}\textsc{Human} & 4.31 &  1.68 & 5.34 \\
    \hspace{2mm}\textsc{MT-Re} & -15.83 & -5.55 & -4.66  \\
    \hspace{2mm}\textsc{MT-Ra} & 4.34 &  2.12 & 2.21 \\
    \hspace{2mm}\textsc{ST-Re} & 0.55 &  1.19 & 1.35 \\
    \hspace{2mm}\textsc{ST-Ra} & 6.74 &  4.93 & 5.14 \\
    \midrule
    CQA \\
    \hspace{2mm}\textsc{Human} & 14.73 & 15.46 &  16.16 \\
    \hspace{2mm}\textsc{MT-Re} & -7.07 & -5.38 &  -3.53 \\
    \hspace{2mm}\textsc{MT-Ra} & -1.31 & 0.32 & 6.33\\
    \hspace{2mm}\textsc{ST-Re} & 3.76 & 1.82 &  2.46 \\
    \hspace{2mm}\textsc{ST-Ra} & 10.32 & 7.24 & 13.43\\
\bottomrule
 \end{tabular}
\caption{We check LAS scores across three random seeds, since random seeds tend to have a large influence on all statistics derived from pretrained neural language models \cite{dodge2020}. Seed 1 is the result reported in the main body. We test two additional seeds for our primary experiments, retraining all models involved in the LAS score (including task model, simulator, and ST generators).}
\vspace{-6pt}
\label{table:seed_robustness}
\end{table}

\begin{table}[t]
    \centering
    \small
\begin{tabular}{l r r} 
\toprule
& \multicolumn{2}{c}{Model} \\
\cmidrule(lr){2-3}
\hspace{-2mm}Method & T5-Base & RoBERTa-Large \\ 
\midrule
    \hspace{2mm}\textsc{Human} & 4.31 (1.97) & -1.09 (2.69) \\
    \hspace{2mm}\textsc{ST-Re} & 0.55 (0.87) & -0.44 (0.95) \\
    \hspace{2mm}\textsc{ST-Ra} & 6.74 (4.53) & 4.74 (9.68) \\
\bottomrule
 \end{tabular}
\caption{LAS score comparison between T5-Base and RoBERTa-Large models with SNLI data (95\% confidence intervals obtained by bootstrap). For ST models, the task model and simulator are of the same architecture. RoBERTa produces lower LAS scores than T5, and their rank ordering is not necessarily the same. The differences between them could result from their pretraining procedures, architectural differences, finetuning sample efficiency, or another cause.}
\vspace{-6pt}
\label{table:model_robustness}
\end{table}

\begin{table}[t]
    \centering
    \small
\begin{tabular}{l @{\extracolsep{\fill}} l c c} 
\toprule
& & \multicolumn{2}{c}{SNLI} \\
\cmidrule(lr){3-4}
Method & & Dev. Acc (CI) & Test Acc (CI) \\ 
\midrule
    \textsc{T5-Base} & & 91.31 (.56) & 91.01 (.57) \\
    \textsc{MT-Re} & & 91.62 (.55) & 91.14 (.56) \\
    \textsc{MT-Ra} & & 91.56 (.55) & 91.20 (.56)  \\
\bottomrule
 \end{tabular}
\caption{NLI results using the full training dataset. Generative models of explanations can maintain task accuracy.}
\label{table:nli_full}
\end{table}

\begin{figure*}
\centering
\includegraphics[width=.98\textwidth]{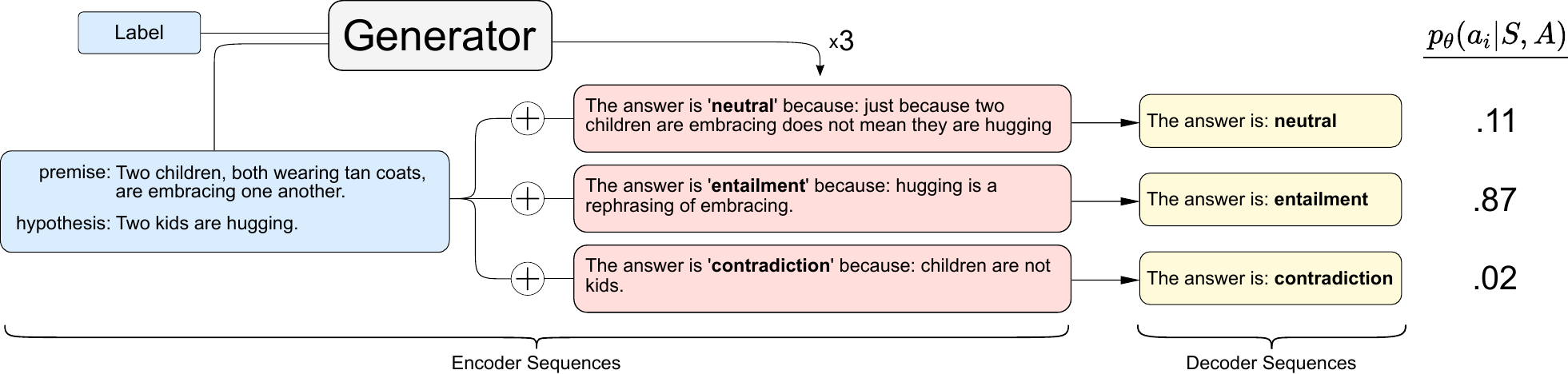}
\caption{Inputs and outputs for the sequence to sequence ST-Ra framework. One explanation is generated for each answer choice, conditioned on the choice. The sequences and answers are supplied to a sequence-to-sequence task model for scoring. We use separate T5 models for the generator and task model.} 
\label{fig:st_ra}
\end{figure*}

\section{LAS Robustness Checks} 
\label{appendix:ovr_robustness}

\subsection{Continuous Leakage Scores and LAS Metric}
\label{appendix:continuous-las}
While we binarize our proxy for label leakage based on prediction correctness and take the raw average of explanation effects across two leakage bins, a continuous measure of leakage can be obtained directly from $p(\hat{y}|\hat{e})$. Then, an arbitrary number of bins can be used. Interestingly, for a T5 model fine-tuned by decoder sequence likelihood maximization, these probabilities are tightly concentrated around values just above random chance performance (.33 for both CQA v1.0 and SNLI), taking a roughly normal distribution. As a result, they are easily calibrated via Platt scaling \cite{platt}. To check for our results' robustness, we perform sensitivity analysis with respect to the number of evenly spaced leakage bins chosen to subset, after calibrating our leakage probabilities. Across bin counts between 2 and 100, LAS estimates typically vary by less than 1 point, and as a result, method ranking is almost always preserved. In the limit of the number of bins, our metric becomes the integral of the explanation effect as a function of leakage probability. To ensure the robustness of LAS scores, this type of sensitivity analysis should be performed whenever possible, but especially when explanation effectiveness is not linearly related to the leakage probability.

\subsection{Robustness to Seed and Model Choice}
\label{appendix:robustness_seed_model}
We check LAS scores across three random seeds since random seeds tend to have a large influence on all statistics derived from pretrained neural language models \cite{dodge2020}. Results are shown in Table \ref{table:seed_robustness}. The rank ordering of scores is typically preserved, and in most cases, scores display relatively low variance, although there are some outlying values. 

We also check the effect of using a different simulator model, shown in Table \ref{table:model_robustness}. We compare between our primary choice of T5-Base and RoBERTa-Large models for SNLI data. For ST models, the task model and simulator are of the same architecture, but we do not evaluate MT conditions since RoBERTa is not generative. RoBERTa produces lower LAS scores than T5, and their rank ordering is not necessarily the same, though \textsc{ST-Ra} is the highest on average in both cases. The differences between them could result from their pretraining procedures, architectural differences, finetuning sample efficiency, or another cause.

\section{Alternative Computational Models and Language Modeling Objectives}
\label{appendix:alternative}
Our generative models neither gained nor lost accuracy relative to their baselines when implemented with T5 models. Since learning from explanations to improve accuracy is another goal in collecting human explanations as data, we seek to assess this trend with alternative computational models and language modeling objectives. Hence, we test our MT models with Masked Language Modeling (MLM) objectives in place of the Causal objectives used for the generation, and wherever a generator or task model appears in current experiments, we test the effect of substituting GPT2 and BERT in their place. We show results for these models in Table \ref{table:acc_alternative_results}; GPT2+BERT methods are tagged as \textsc{Enc} methods. Just as with our generative approaches, we observe no differences in accuracies between baselines and other methods.

\begin{table}[t]
    \centering
    \small
\begin{tabular}{c @{\extracolsep{\fill}} l c c c} 
\toprule
\hspace{-17pt}\textbf{LAS} & & \multicolumn{3}{c}{Human} \\
\cmidrule(lr){3-5}
 Model & & -1 & 0 & 1 \\ 
\midrule
     -1 && 0.271 & 0.659 &0.071 \\
      0 &&  0.082& 0.781& 0.138 \\
      1 &&  0.031& 0.654& 0.315 \\
\bottomrule
\end{tabular}
\caption{Row-normalized contingency table between model-based and human variables resulting from the expert simulation analysis. Model scores of -1 and 1 tend to shrink toward human ratings of 0.}
\label{table:sim_validation_norm}
\end{table}

\begin{table}[t]
    \centering
    \small
\begin{tabular}{l @{\extracolsep{\fill}} l c c} 
\toprule
& & \multicolumn{1}{c}{e-SNLI} & \multicolumn{1}{c}{\textsc{CQA}}  \\
\cmidrule(lr){3-3} \cmidrule(lr){4-4}
Method & & Test Acc (CI) & Dev Acc (CI) \\ 
\midrule
    \textsc{BERT-Base} & & 87.01 (0.66) & 67.89 (2.97) \\
    \textsc{ST-Re-Enc} & & 85.67 (0.69) & 63.16 (3.07) \\
    \textsc{ST-Ra-Enc} & & 85.62 (0.69) & 64.84 (3.04) \\
    \textsc{MT-Re-Enc} & & 87.25 (0.66) & 70.74 (2.89) \\
    \textsc{MT-Ra-Enc} & & 87.23 (0.66) & 69.79 (2.92) \\
    \addlinespace
    \textsc{T5-Base} & & 88.14 (0.63) & 68.84 (2.95)  \\
    \textsc{MT-Re-MLM} & & 88.26 (0.63) & 69.05 (2.94) \\
    \textsc{MT-Ra-MLM} & & 88.43 (0.63) & 70.11 (2.91) \\
\bottomrule
 \end{tabular}
\caption{Task results table with alternative computational models and language modeling objectives. }
\label{table:acc_alternative_results}
\end{table}

\section{Human Quality Rating Collection}
\label{appendix:mturk}
\begin{figure*}
\centering
\includegraphics[width=.8\textwidth]{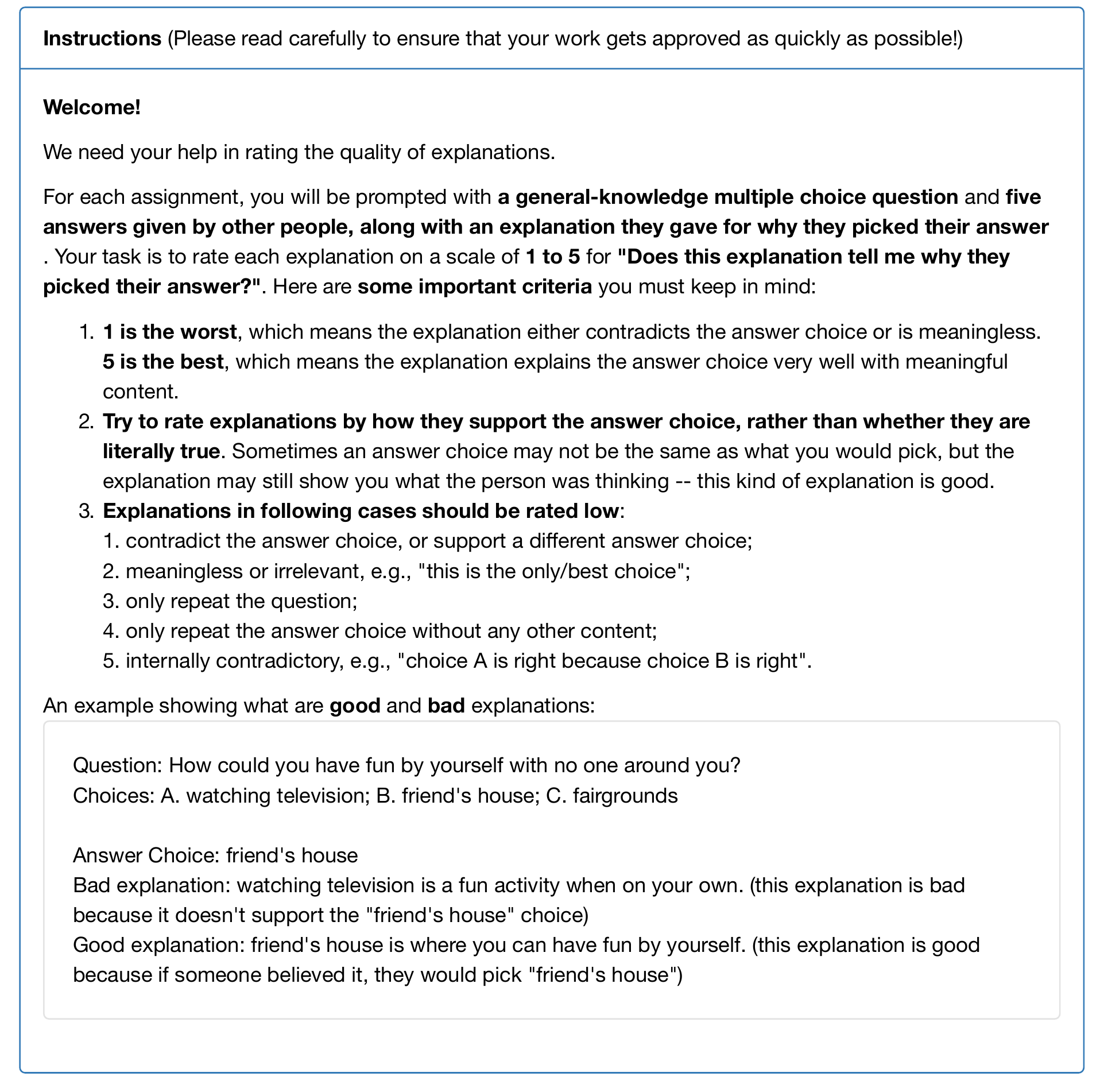}
\caption{The instruction shown on Amazon Mechanical Turk page for human rating collection on CQA.} 
\label{fig:mturk_ins}
\end{figure*}

\begin{figure*}
\centering
\includegraphics[width=.8\textwidth]{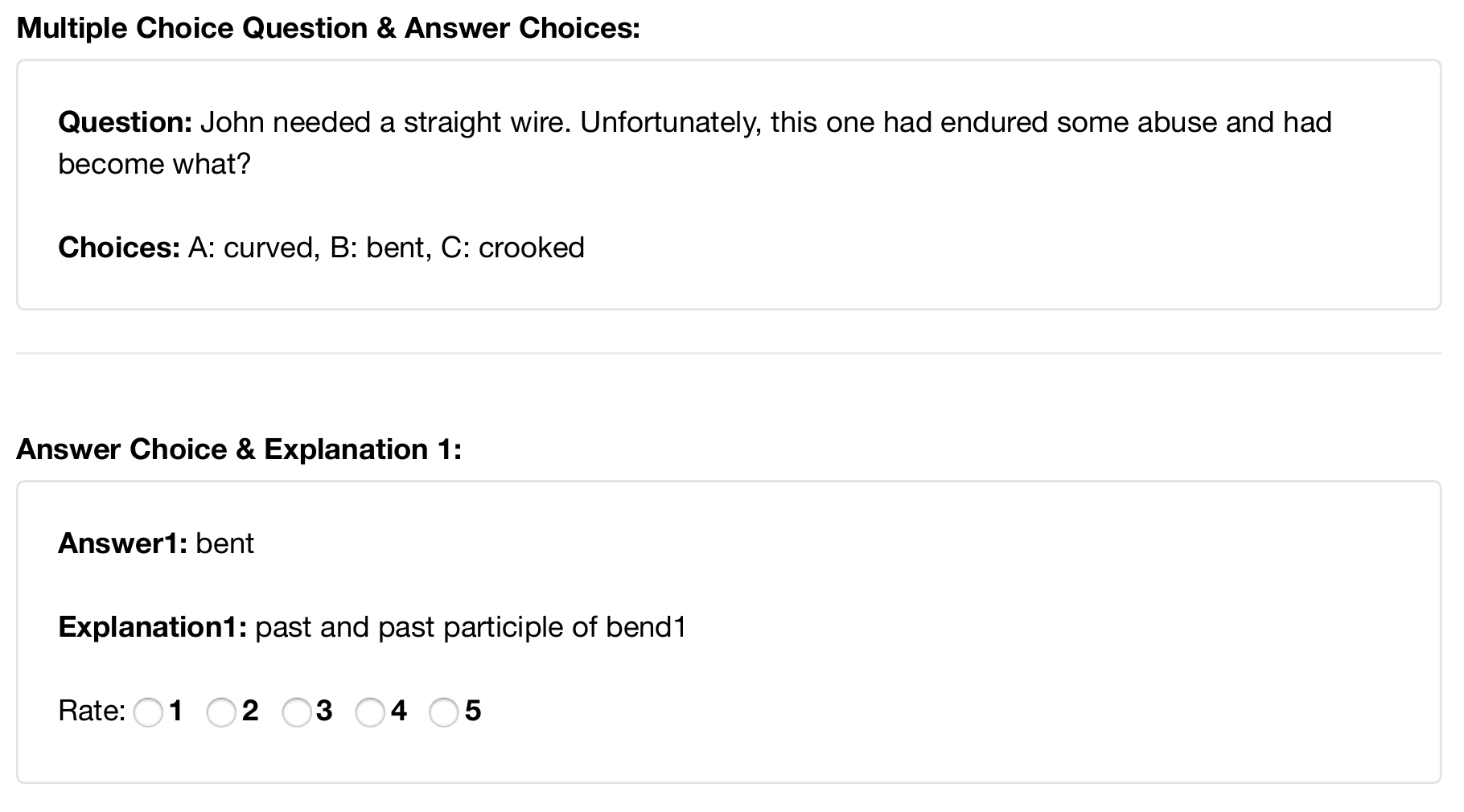}
\caption{A part of the questions for human rating collection on CQA.} 
\label{fig:mturk_que}
\end{figure*}

We collected the human ratings of explanation quality from Amazon Mechanical Turk. For CQA or SNLI, we sample 200 examples from the development or testing set (CQA's testing set does not contain human explanations). Each example has five explanations that are generated by the four models we introduced in the main paper as well as humans. We anonymously shuffle the five explanations and ask turkers to rate them separately on a 5-point Likert scale. Meanwhile, we give them some instructions about ``rate explanations by how they support the answer choice, rather than whether they are literally true'' and ``explanations in which cases should be rated low''. Figure~\ref{fig:mturk_ins}
shows the full instructions we used for collecting explanation ratings for CQA, and Figure~\ref{fig:mturk_que} shows one CQA question and its answer choices plus the first model's choice and its explanation. SNLI has a similar GUIs. Turkers will be required to rate five (choice, explanation) pairs on one page.

We collected 3 responses for each example, so there are 600 responses in total for each dataset. We apply a simply quality filter to filter the responses from bad turkers. We first manually picked 10 explanations from both CQA and SNLI that contradict their corresponding model outputs (choices). As we know, these explanations are sure to be bad. So, we filter the responses from those turkers who rated high ($>2$ for CQA, $>3$ for SNLI, since SNLI has a higher average rating) for these bad explanations. After filtering, we finally obtained 466 responses for CQA and 436 responses for SNLI.

